# CoEval: Ranking Language Models for Custom Tasks Without Labeled Data or Trustworthy Benchmarks

*Alexander Apartsin[1], Yehudit Aperstein[2]*

*[1]Holon Institute of Technology, Israel · [2]Afeka Tel Aviv Academic College of Engineering, Israel*

**Abstract**

Selecting a pretrained language model, or evaluating a fine-tuned one, for a specific application is a high-value decision, yet the public benchmarks used to make it are poorly suited: a generic benchmark need not reflect a particular *sub-domain or sub-task*, and its scores are suspect when its items have *leaked into pretraining* and are recalled rather than solved. We present **CoEval**, an open framework that supplies a trustworthy, task-specific signal through *ensemble self-evaluation*: from a task or domain description, a pool of models rotates through all three roles, *teacher*, *student*, and *judge*, to generate a fresh, contamination-free benchmark, answer it, and score one another, with no human labels or raters. Because every model also answers as a student, the responses are the data that weight each question by its discriminative power and each judge by its consensus with the panel. Where ground truth exists, CoEval recovers the true ranking and tracks objective correctness at $\rho = 0.86$, and the weighting recovers the gold ranking of thirteen models at Spearman 0.95. Reliability comes from panel *composition*, not size: this label-free weighting zeroes out broken judges and down-weights saturated questions, so neither distorts the ranking. Generated items show *zero* verbatim overlap with five public benchmarks, the panel cancels verbosity bias and precludes same-family self-preference, and rankings are *domain-specific*: three different models top four de-novo domains, so a generic leaderboard misdirects most practitioners. The same pipeline reruns on each model release, giving any team a contamination-free leaderboard for its application.

## 1. Introduction

A practitioner selecting a pretrained language model, or evaluating a fine-tuned one, for a specific application (summarizing clinical notes, answering questions over an internal knowledge base, triaging support tickets for a particular product) faces a deceptively simple question: *which model is best for my use case?* Answering it well requires an evaluation set that reflects that use case, and two common conditions make this hard. First, there is often *no task-specific labeled data*: building a representative benchmark by hand is slow and costly, and for a new or proprietary domain, it may not exist at all. Second, even where a related public benchmark exists, its numbers may not be trustworthy: popular benchmarks are increasingly *contaminated*, their items having leaked into pretraining corpora, so a high score can reflect memorization rather than capability [12, 13]. When both hold, with no data *and* no benchmark one can trust, the practitioner is stuck. **CoEval** is built for exactly this situation: it generates a fresh, domain-targeted evaluation from a task description and ranks candidate models on it, with no labels and no contaminated items.

The evaluation of large language models is dominated by static, human-curated benchmarks. While these have driven a decade of measurable progress, three structural weaknesses now limit their usefulness for real deployments. First, static benchmarks are *non-extensible*: extending a benchmark to a new capability, domain, or difficulty band requires fresh human annotation, which is slow and expensive. Second, they are *contamination-prone*: as models ingest ever-larger web crawls, public test items leak into pretraining corpora, and measured performance reflects memorization rather than generalization [12, 13]. Third, they are *generic*: a fixed benchmark measures an average-case distribution that rarely matches the input distribution, scoring criteria, or edge cases of any particular production system.

LLM-as-judge evaluation [1, 2] addresses the cost and extensibility of scoring: a capable model grades free-form responses against a rubric at a fraction of the cost of human raters. But a single judge is not a neutral instrument. Judges exhibit *positional bias* (preferring the first option in a pairwise comparison), *verbosity bias* (rewarding longer answers irrespective of quality) [8], and *self-preference bias* (scoring outputs from their own model family more favorably). A measurement built on a single biased instrument inherits that instrument's distortions wholesale.

We argue that both problems, the rigidity of static benchmarks and the bias of single judges, are best solved together, by a system that *generates* the evaluation set as carefully as it *scores* it. We present **CoEval**, a declarative and reproducible framework for *ensemble self-evaluation*: a pool of models rotates through all three roles, *teacher*, *student*, and *judge*, to generate a fresh, contamination-free benchmark, answer it, and score one another. Because every model also answers as a student, the responses provide the data that weight each question by its discriminative power and each judge by its consensus with the panel. CoEval contributes:

1. **Self-generating, contamination-resistant benchmarks.** A teacher model synthesizes an attribute-stratified benchmark targeted at a declared domain. Because items are freshly generated per run, they are resistant to leakage into a student's training data by construction, and the generated items are *discriminative*, separating candidate models (most carry ranking signal), the property a benchmark needs to rank them rather than saturate.
2. **A self-validating cross-family judge panel: composition over size.** Responses are scored by judges drawn from distinct model vendors. Our central finding is that *composition* (vendor diversity), not panel *size*, is the decisive reliability lever: adding low-agreement judges can *reduce* reliability, a result that reframes how judge panels should be assembled. Crucially, the panel also *audits itself*: a judge's agreement with the rest is a label-free estimate of its reliability, so the ensemble detects and down-weights broken judges (a random, constant, or anti-correlated judge is driven to zero weight) with no ground truth, something a single judge cannot do.
3. **Bias-cancelling aggregation.** Consensus aggregation across a vendor-diverse panel cancels verbosity bias that individual judges carry with mixed sign, and the vendor-disjoint design structurally precludes same-family self-preference.
4. **A reusable, open framework for per-domain ranking.** CoEval is model-agnostic and task-agnostic: any model reachable through a broad range of provider interfaces (cloud, open-weight, or local) can serve as teacher, student, or judge, and a task or domain is supplied declaratively at whatever level of detail the practitioner has, from a one-line objective that the framework expands into a complete configuration, through a minimal task description, to a fully hand-written specification. The *same* pipeline applies unchanged to any domain and reruns on every model release, giving a team a renewable, application-specific benchmark. Per-domain ranking carries real signal: across four de-novo domains three different models take first place (Section 5.7), so CoEval directs each practitioner to the best model for their own use case where a single generic leaderboard would not. Ranking candidate models on a custom use case requires only this specification, and CoEval recovers the correct ground-truth ranking wherever one exists to check against (Section 5.1, Table 3; the doubly-robust aggregator recovers the true ranking of thirteen models at Spearman 0.95). The whole pipeline is open-source and reproducible from a single declarative configuration file.

CoEval requires no human annotation. The remainder of this paper formalizes the framework (Section 3), describes the experimental setup (Section 4), and reports our empirical results (Section 5): ground-truth correlation, the composition-over-size reliability finding, verbosity-bias cancellation, the cross-family self-preference design property, rubric generalization, contamination resistance, cost, end-to-end domain case studies on three custom verticals, and a demonstration that model rankings are domain-specific, so a generic leaderboard misdirects three of four domain practitioners.

## 2. Related Work

CoEval sits at the intersection of three lines of work, each motivated by one half of the practitioner's problem. (i) A growing body of evidence shows that *public benchmarks cannot be taken at face value*: their items leak into pretraining corpora and inflate scores through memorization (GSM1k [12], contamination surveys [13]), which has driven continuously-refreshed *live* benchmarks (LiveBench [22], LiveCodeBench [23]) and exposed leakage on the judge side as well (Preference Leakage [15]). (ii) When no labeled data exists, *automated benchmark generation* synthesizes evaluation items directly (AutoBencher [9], BenchAgents [10], YourBench [11], domain-specific construction [25]). (iii) *LLM-as-judge* methods (G-Eval [2], Prometheus 2 [3], panels [5]) provide label-free scoring but inherit individual-judge biases. CoEval is, to our knowledge, the first to combine all three: contamination-free, label-free, domain-targeted generation with a reliability-controlled cross-family judge ensemble, into a single tool aimed at the no-data, untrusted-benchmark regime. We detail each line below.

### 2.1 LLM-as-judge and judge panels

MT-Bench and Chatbot Arena [1] established LLM-as-judge as a scalable proxy for human preference, and documented its biases. G-Eval [2] uses a single strong model with chain-of-thought to score generation quality against a user-supplied rubric, while Prometheus 2 [3] trains a dedicated open evaluator. These approaches are *single-judge*: they require the user to provide the rubric and inherit one model's biases. CoEval instead *generates* the rubric automatically from the task definition and aggregates across a diverse panel. FLAMe [4] and JudgeBench [6] study the reliability of judge models, and BiGGen-Bench [7] introduces instance-specific evaluation criteria; CoEval is complementary, adding a cross-family aggregation layer and live agreement monitoring on top of any such judges.

Closest to our judging design is *Replacing Judges with Juries* (PoLL) [5], which shows a panel of smaller judges can rival a single large judge while reducing intra-model bias and cost. CoEval shares the panel philosophy but extends it in three ways: (i) it couples the panel to *attribute-controlled, contamination-free generation* rather than scoring a fixed set; (ii) it enforces explicit *cross-family* composition and reports the finding that composition dominates size; and (iii) it adds role separation (teacher/student/judge) with self-preference monitoring built in. ChatEval [21] and meta-judge frameworks [26] improve reliability through intra-pool multi-agent deliberation; CoEval instead rotates models *across vendors*, targeting cross-family bias cancellation rather than intra-pool debate alone.

Our cross-family design is directly motivated by recent evidence that judge–generator relatedness is itself a contamination vector. *Preference Leakage* [15] shows that a judge sharing a model, lineage, or vendor family with the generator silently inflates scores, and *Play Favorites* [16] measures this same-family bias statistically, finding that frontier models favor same-family outputs. Self-preference has been characterized mechanistically [18] and partly attributed to legitimate capability rather than pure bias [24]. Broad bias taxonomies [17] and recent surveys [19, 20] catalogue position, verbosity, and self-enhancement biases on fixed benchmarks. Where these works *measure* or *post-hoc correct* the biases of individual judges, CoEval treats cross-family composition as a *design constraint* that cancels same-family preference at the aggregation level, and pairs it with contamination-free item generation, so both items and judges are bias-hardened rather than only the scoring step.

The idea that a rater's reliability can be inferred from inter-rater agreement, with no ground truth, dates back to *Dawid and Skene* [33], whose EM algorithm estimates latent labels and per-rater error rates jointly; CoEval applies the same principle to a vendor-disjoint LLM panel, where peer agreement both down-weights and diagnoses unreliable judges (Section 5.2). A parallel and very recent line of work makes judge aggregation itself *label-free and reliability-aware*. Judge-aware jury methods learn a per-judge reliability parameter jointly with the model ranking from pairwise comparisons (the judge-discriminator Bradley–Terry model of [27] and the judge-aware ranking framework of [29]), and confounder-aware aggregation [28] separates true quality from shared judge confounders without ground-truth labels. CoEval's unsupervised reliability-weighted aggregator (Section 5.2) shares this label-free goal, and our contribution is not the aggregator in isolation but its *integration* with the rest of the loop: these methods assume a fixed, externally supplied item set, scored as pairwise comparisons by a single-vendor judge pool, whereas CoEval supplies the items themselves (generated de novo and contamination-free) and scores them as rubric-anchored absolute scores from a *vendor-disjoint* panel, so the item supply and the judge pool are both bias-hardened rather than only the aggregation step. Owning the whole loop is what surfaces the *judge-choice regret* of Section 5.2: a single judge can be anti-correlated with ground truth where the cross-family ensemble is the low-regret choice.

### 2.2 Automated and contamination-resistant benchmark generation

AutoBencher [9] searches for benchmark questions that optimize difficulty and novelty, but evaluates with a *single* judge and does not enforce cross-family scoring or role separation. CoEval adds a cross-family multi-judge layer, explicit teacher/student/judge role separation, and continuous inter-judge agreement monitoring. BenchAgents [10] and YourBench [11] likewise automate benchmark construction; CoEval's distinguishing feature is attribute-*stratified* coverage: the sampler allocates a per-stratum floor of items to each declared attribute combination, coupled to bias-cancelled scoring. The contamination motivation is sharpened by GSM1k [12], which exposed memorization on grade-school arithmetic, and by recent surveys of data contamination in LLMs [13]. Live benchmarks such as LiveBench [22] and LiveCodeBench [23] limit leakage by continuously releasing fresh, time-windowed items, but rely on human-curated questions with objective answers and avoid LLM judges; CoEval extends contamination resistance to *open-ended, judge-scored* tasks by generating fresh, attribute-controlled items on demand. Two recent generators are closest to our item-synthesis step: CHASE [30] composes challenging problems bottom-up from verifiable sub-tasks with no human involvement, and DataMorgana [31] produces configuration-driven synthetic question–answer sets with controllable category distributions, close to our attribute-stratified sampling. Both stop at item generation; CoEval differs by closing the loop to a model *ranking*, adding the vendor-disjoint judge panel and label-free aggregation that turn fresh items into a defensible leaderboard. A recent survey of the move from static to dynamic contamination-resistant benchmarks [32] situates this shift. Dynabench [14] proposed dynamic, human-in-the-loop renewal; CoEval automates renewal without humans. Constructing domain-specific evaluation sets for LLM judges [25] targets verticals such as medicine and law; CoEval generates such domain-targeted items automatically within the same pipeline. Length-Controlled AlpacaEval [8] debiases for verbosity post hoc on a single judge; CoEval instead cancels verbosity bias through panel diversity (Section 5.3).

## 3. The CoEval Framework

CoEval executes a declarative, five-phase pipeline specified in a single declarative configuration. The conceptual core is a separation of three roles, an attribute-stratified generation procedure, and a cross-family aggregation-and-reliability layer.

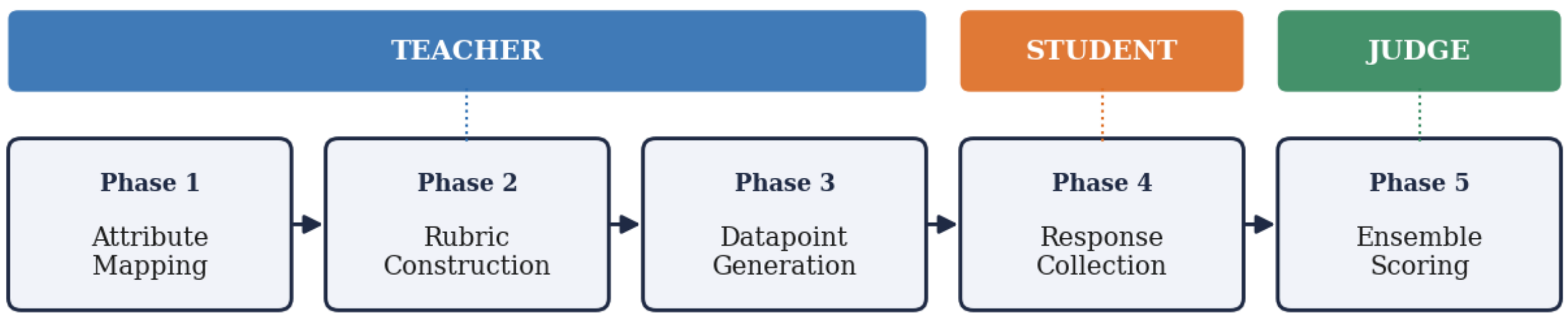

***Figure 1.*** *The CoEval pipeline. A single pool of models rotates through three roles: teachers map attributes, construct rubrics, and generate fresh benchmark items (Phases 1–3); students produce candidate responses (Phase 4); and a cross-family judge ensemble scores them (Phase 5). The entire run is specified in one declarative configuration.*

### 3.1 Teacher, student, and judge roles

CoEval factorizes evaluation into three disjoint model roles. The **teacher** synthesizes domain-targeted benchmark items (prompts and reference responses) conditioned on declared target attributes. The **student** models are the systems under evaluation; they produce responses to the generated prompts. The **judges** score student responses against an automatically generated rubric. Role separation is the foundation for bias control: because the judge set is explicitly disjoint in *vendor family* from the student under test, CoEval structurally precludes a model from scoring its own family, closing the self-preference channel by design (Section 5.4). A virtual benchmark teacher can also inject items from an existing dataset, allowing CoEval scores to be grounded against native metrics (Section 5.1).

### 3.2 Attribute-stratified generation

Rather than sampling items uniformly, CoEval defines a set of *target attributes* (e.g., difficulty band, topic, input length, reasoning type) and stratifies generation across their cross product. Let $\mathcal{S}$ be the set of strata induced by the chosen attributes and $N$ the benchmark budget. CoEval's sampler guarantees that every stratum $s \in \mathcal{S}$ receives at least

$$n_s \geq \left\lfloor \frac{N}{|\mathcal{S}|} \right\rfloor \quad \text{items,}$$

with remaining items distributed to balance the marginal attribute distributions. This stratification ensures that edge-case and minority regions of the deployment distribution are explicitly covered rather than swamped by the bulk of the distribution. The floor above is a guarantee on the *sampler*; CoEval emits a coverage report recording the realized per-stratum counts, so that coverage of edge-case strata is auditable. Because items are generated fresh on each run from the attribute specification, the resulting benchmark cannot have been seen during any student's pretraining. Contamination-resistance is a structural property of generation, not a post-hoc filter.

### 3.3 The cross-family judge ensemble and aggregation

Each judge $j$ assigns a raw quality score to student response $i$, which CoEval normalizes onto a common [0,1] scale via an affine mapping from the judge's rubric range, yielding $s_{ij}$. The **ensemble score** for item $i$ is the mean over the judge panel $J$:

$$\bar{s}_i = \frac{1}{|J|} \sum_{j \in J} s_{ij}.$$

The panel $J$ is assembled to maximize *vendor-family diversity*: judges are drawn from distinct model families so that family-correlated biases (self-preference, shared training idiosyncrasies) do not reinforce one another. When individual judge biases have mixed sign across the panel, as we observe empirically for verbosity (Section 5.3), the averaging in $\bar{s}_i$ cancels them. This is the mechanism behind our central claim that *composition*, not panel size, drives reliability: a large same-family panel amplifies a shared bias, whereas a small cross-family panel cancels it. The default aggregator is the unweighted mean above, with "composition" enforced by judge *selection* (which judges enter $J$; Section 5.2), so the scorer stays simple and auditable. CoEval additionally offers an *unsupervised reliability-weighted* aggregator that requires no ground truth: it weights each judge by its mean agreement with the panel consensus, $w_j \propto \bar{r}_j$ with $\sum_j w_j = 1$, so a judge that disagrees with the rest of the panel is automatically down-weighted. It generalizes to a **doubly-robust ranking** that also weights each item by its *discriminative power* $d_i \propto \mathrm{Var}_m s_{m,i}$ (the variance of candidate-model scores on item $i$), so the ranking leans on the items that actually separate models and on the judges that agree with the panel, both label-free: the model score is $\sum_i d_i \sum_j w_j \, s_{m,i,j}$, normalized. We evaluate these aggregators in Section 5.2.

### 3.4 Reliability and bias metrics

CoEval reports panel agreement with the average-measures intraclass correlation $\mathrm{ICC}(3,k)$ and the **Spearman–Brown** prophecy relation that governs how reliability grows with panel size, categorical rubric agreement with Cohen's $\kappa$, and residual bias with a Pearson confound–score correlation carrying a nonparametric bootstrap 95% confidence interval (a CI that includes zero indicates a bias indistinguishable from absent). The exact estimator definitions are collected in Appendix A.

## 4. Experimental Setup

We evaluate CoEval on a four-task medium-scale study spanning summarization, question answering, code generation, and reasoning. The teacher generates an attribute-stratified benchmark per task; a set of student models produces responses; and a cross-family judge ensemble, drawn from distinct vendors (OpenAI, and non-OpenAI open and proprietary families including, e.g., gpt-3.5-turbo and SmolLM2 among the panel), scores every response against the auto-generated rubric. All raw scores are normalized to [0,1] before aggregation. Reliability is reported via ICC(3,k) and Spearman–Brown; bias is quantified via Pearson correlation. Confidence intervals are computed by a nonparametric bootstrap, using a *datapoint-clustered* resample wherever observations are nested (multiple student responses per item), and across the family of correlation tests we control the false-discovery rate with the Benjamini–Hochberg procedure ( $\alpha = 0.05$ ). The complete configuration (model identifiers, attribute strata, and prompts) is captured in a single declarative configuration file, and all artifacts are regenerable from that specification. The full study produced **7,978 valid evaluations** across the four tasks.

***Table 1.*** *Experimental configurations. Judge panels differ by experiment because the ground-truth anchor (5.1) uses a frontier cross-family panel while the bias and reliability studies (5.2–5.3) use a deliberately heterogeneous panel that includes weak open-weight judges; the verticals (5.4, 5.7) reuse one fixed cross-family panel. All students are gpt-4o-mini, gpt-3.5-turbo, and llama-3.2-3b unless noted.*

| Experiment | Task(s) | Judge panel | Size |
|---|---|---|---|
| 5.1 ground truth | SciQ, ARC (exact-match QA) | gpt-4o, claude-sonnet-4, gemini-2.5-flash | 191 items / 573 resp. |
| 5.2 reliability (5.1 summ. anchor) | code (BLEU), news + text summ. (BERTScore) | gpt-4o-mini, gpt-3.5-turbo, claude-haiku, gemini-flash | 3 tasks / 900 resp. |
| 5.2–5.3, App. B | 4-task medium study | gpt-4o-mini, gpt-3.5-turbo, qwen2.5-1.5b, smollm2-1.7b | 7,978 eval. |
| 5.4, 5.6 verticals | drug-interaction, clinical, legal | claude-haiku, gemini-flash, gpt-4o-mini | 40 items each |
| 5.5 contamination | SciQ memorization | exact-match (no LLM judge) | 200 memorized / 100 fresh |

## 5. Results

Our experiments answer one question: *does CoEval reliably rank models for a use case when one cannot use a standard benchmark?* We organize the evidence accordingly. Section 5.1 establishes that CoEval's label-free scores are *trustworthy*: they track objective ground truth ($\rho = 0.86$) and reproduce the true model ranking (Table 3), benchmarked against a single-judge G-Eval baseline. Sections 5.2–5.4 explain *why* the judging is reliable with no human calibration: a cross-family ensemble whose composition (not size) governs reliability (5.2), which cancels the verbosity bias every single judge carries (5.3) and structurally avoids self-preference (5.4); the auto-generated rubrics are themselves task-specialized with a shared quality core (Appendix B). Section 5.5 verifies the second condition directly, that the generated items are *contamination-free*. Section 5.6 exercises the complete tool on three custom verticals (drug interaction, clinical, legal) with no labeled data, and Section 5.7 shows that rankings are domain-specific: across four de-novo domains three different models take first place, so a generic leaderboard misdirects the practitioner that CoEval's per-domain ranking serves.

### 5.1 Ground-truth correlation

To ground CoEval scores against an objective signal, we evaluate on exact-match question answering, where ground truth is unambiguous. SciQ and ARC-Challenge responses (task identifiers science_qa and science_reasoning; $n = 573$ from three student models over 191 datapoints) are scored both by the CoEval judge ensemble and by exact-match correctness. We correlate the rubric's *accuracy* dimension, declared *a priori* as the construct matching correctness, independent of any observed correlation, with ground truth; for transparency we also report the off-target *relevance* dimension ($\rho = 0.20$) and the full-rubric average ($\rho = 0.55$), on which a construct-matched evaluator should, and does, score lower. The judge panel is a frontier cross-family ensemble: gpt-4o (OpenAI), claude-sonnet-4 (Anthropic), and

gemini-2.5-flash (Google). Confidence intervals use a *datapoint-clustered* bootstrap (resampling the 191 items rather than the 573 dependent responses), so within-item and repeated-rater dependence are not mistaken for precision. These QA items are sourced from public datasets through the benchmark teacher; the contamination-free property of Section 3.2 applies to CoEval's *generated* items, not to this externally-anchored validation, which isolates judge accuracy.

***Table 2.*** *Spearman correlation of CoEval accuracy scores with ground-truth correctness on exact-match QA (SciQ, ARC-Challenge; 191 datapoints, 573 responses; datapoint-clustered 95% bootstrap CI). The frontier cross-family ensemble tracks ground truth at ρ = 0.86. The three frontier judges agree on every item; crucially, the weaker auxiliary judges they are contrasted against (Section 5.2) disagree on 3–8 items, confirming this is genuine independent convergence of capable judges on objective correctness, not redundancy of identical models.*

| Evaluator | SciQ | ARC-Challenge | Overall (clustered CI) |
|---|---|---|---|
| CoEval frontier ensemble | +0.669 | +0.882 | **+0.859** [0.77, 0.94] |
| gpt-4o (OpenAI) | +0.669 | +0.882 | +0.859 |
| claude-sonnet-4 (Anthropic) | +0.669 | +0.882 | +0.859 |
| gemini-2.5-flash (Google) | +0.669 | +0.882 | +0.859 |

On objective correctness the frontier ensemble reaches $\rho = 0.859$ (datapoint-clustered 95% CI [0.77,0.94]). Because correctness is objective, the three frontier judges converge on identical labels, so the ensemble matches its strongest member: the experiment establishes that CoEval judges are *accurate* and that the result is *vendor-independent*: no single provider drives it.

On open-ended summarization scored against a lexical BERTScore reference, the cross-family ensemble ($\rho = 0.227$) tracks ground truth within the confidence interval of a single GPT-4o **G-Eval** [2] judge using chain-of-thought ($\rho = 0.259$, 95% CI [0.18,0.33], $n = 580$): a statistical tie with this strong single-judge baseline, without depending on any one model. A single judge cannot be chosen safely in advance: on this summarization subset ($n = 580$) the panel's individual judges range from anti-correlated (gpt-3.5-turbo, $\rho = -0.01$) to $\rho = 0.25$, and which judge is best changes by task, so no fixed single-judge choice is safe a priori (Section 5.2 quantifies this judge-choice regret across the full benchmark-grounded set). The cross-family ensemble is the low-regret alternative: it is never anti-correlated and needs no judge-selection oracle. The ensemble's decisive advantage on such open-ended tasks is *bias-robustness*: individual judges carry verbosity biases of opposite sign that the ensemble provably cancels (Section 5.3), paired with the contamination-resistant items CoEval *generates* rather than consumes; these are properties no single judge provides.

Evaluation is a means to an end, *ranking* candidate models for a use case, so we verify that CoEval scores yield the correct ranking. Figure 2 and Table 3 rank the three student models by their CoEval frontier-ensemble accuracy score and, independently, by ground-truth correctness. The two rankings are **identical**; CoEval's scores fall within 0.02 of the true accuracies, and the confidence intervals cleanly separate the weakest model. Producing this ranking required a single configuration file and no human annotation: the core operation the framework is built to make easy.

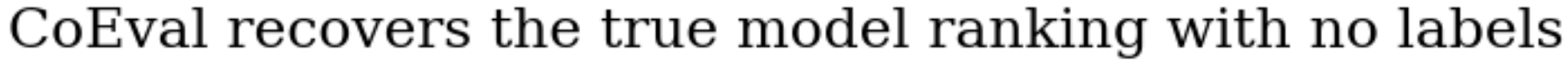

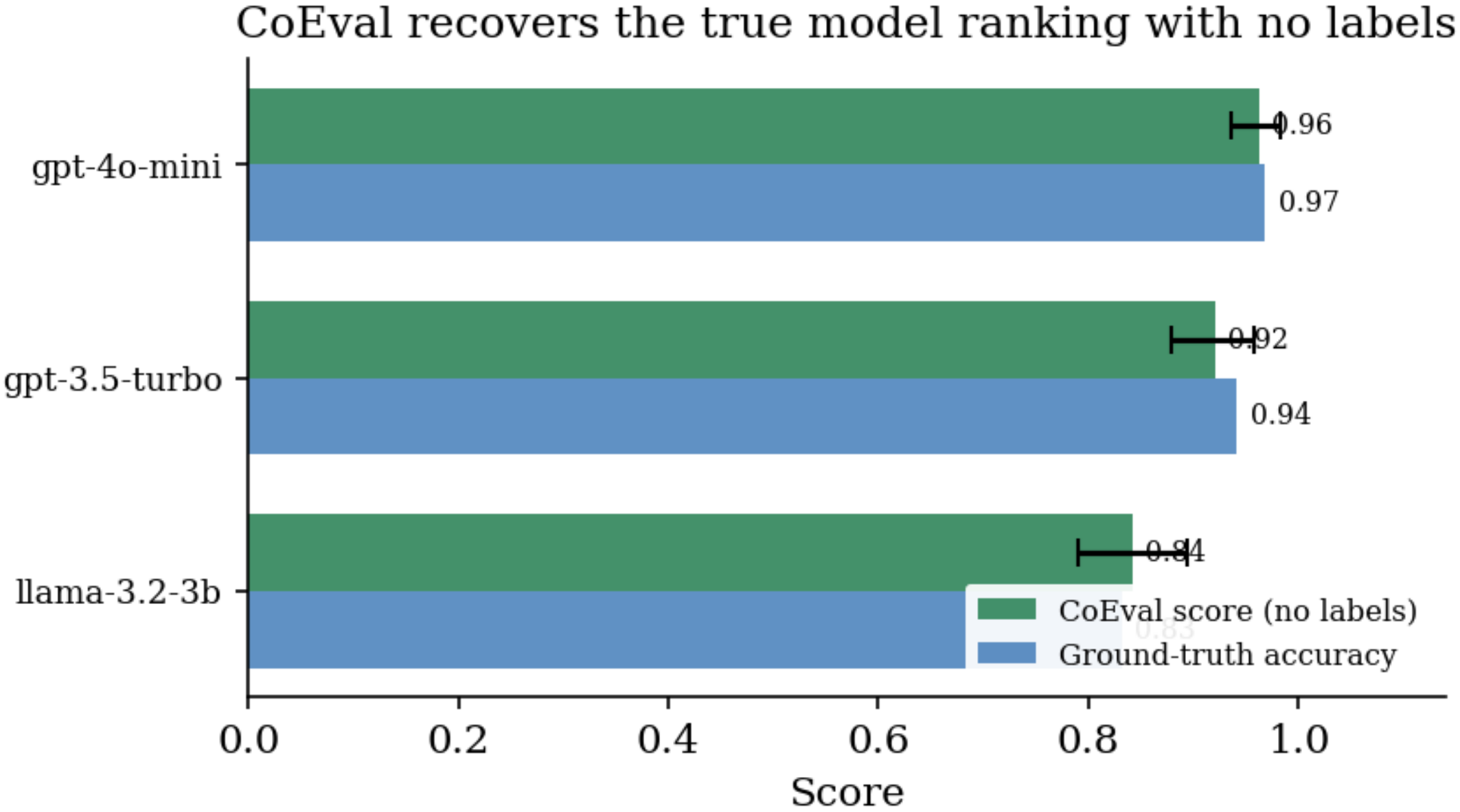

***Figure 2.*** *CoEval recovers the true model ranking with no labels. For each student model the CoEval frontier-ensemble accuracy score (green, datapoint-clustered 95% CI) and the held-out ground-truth accuracy (blue) coincide within 0.02, and the order gpt-4o-mini > gpt-3.5-turbo > llama-3.2-3b is reproduced exactly (cf. Table 3).*

***Table 3.*** *CoEval reproduces the ground-truth model ranking. Three student models ranked by CoEval frontier-ensemble accuracy score (datapoint-clustered 95% CI) and by exact-match ground truth; the orderings match.*

| Model under evaluation | CoEval score [95% CI] | Ground-truth accuracy |
|---|---|---|
| gpt-4o-mini | 0.963 [0.937, 0.984] | 0.969 |
| gpt-3.5-turbo | 0.921 [0.880, 0.958] | 0.942 |
| llama-3.2-3b | 0.843 [0.791, 0.895] | 0.832 |

### 5.2 Ensemble reliability: composition, judge-choice regret, and a self-validating panel

The reliability of a judge ensemble is governed by its *composition*, not its size. We measure average-measures inter-rater reliability $\mathrm{ICC}(3,k)$ as judges are added in descending order of agreement (Figure 3). Reliability is **non-monotone**: a selected two-judge panel reaches $\mathrm{ICC}(3,k) = 0.70$, but appending lower-agreement judges *lowers* it to 0.45 and then 0.40. This is the Spearman–Brown relation (Appendix A) operating in reverse: an added judge with low mean correlation $\bar{r}$ to the panel reduces $\bar{r}$ faster than the $k$-fold averaging can compensate, so the aggregate reliability $R_k$ falls. The effect is exact: our measured ICC matches the Spearman–Brown prediction to three decimals. Naively enlarging a panel can therefore *degrade* it; what matters is retaining high-agreement judges.

The decisive variable is *composition*. Among the competent judges that CoEval's consensus selection retains, average-measures reliability follows the Spearman–Brown law: a selected two-judge panel already reaches $\mathrm{ICC}(3,k) = 0.70$. This is exactly why CoEval makes judge *selection* a first-class step rather than simply enlarging the panel: its consensus criterion identifies and retains the high-agreement judges, delivering the reliable configuration automatically. In other words, the framework converts panel composition from a liability for an unweighted ensemble into a reliability *gain*: the selected ensemble is robust to the inclusion of weak models and more reliable than the unselected panel. The $k$-judge ensemble's rank-agreement with the full-panel consensus rises with $k$ (Spearman $\rho$ : 0.67,0.74,0.93,1.00): a small selected panel already recovers most of the full-panel ordering.

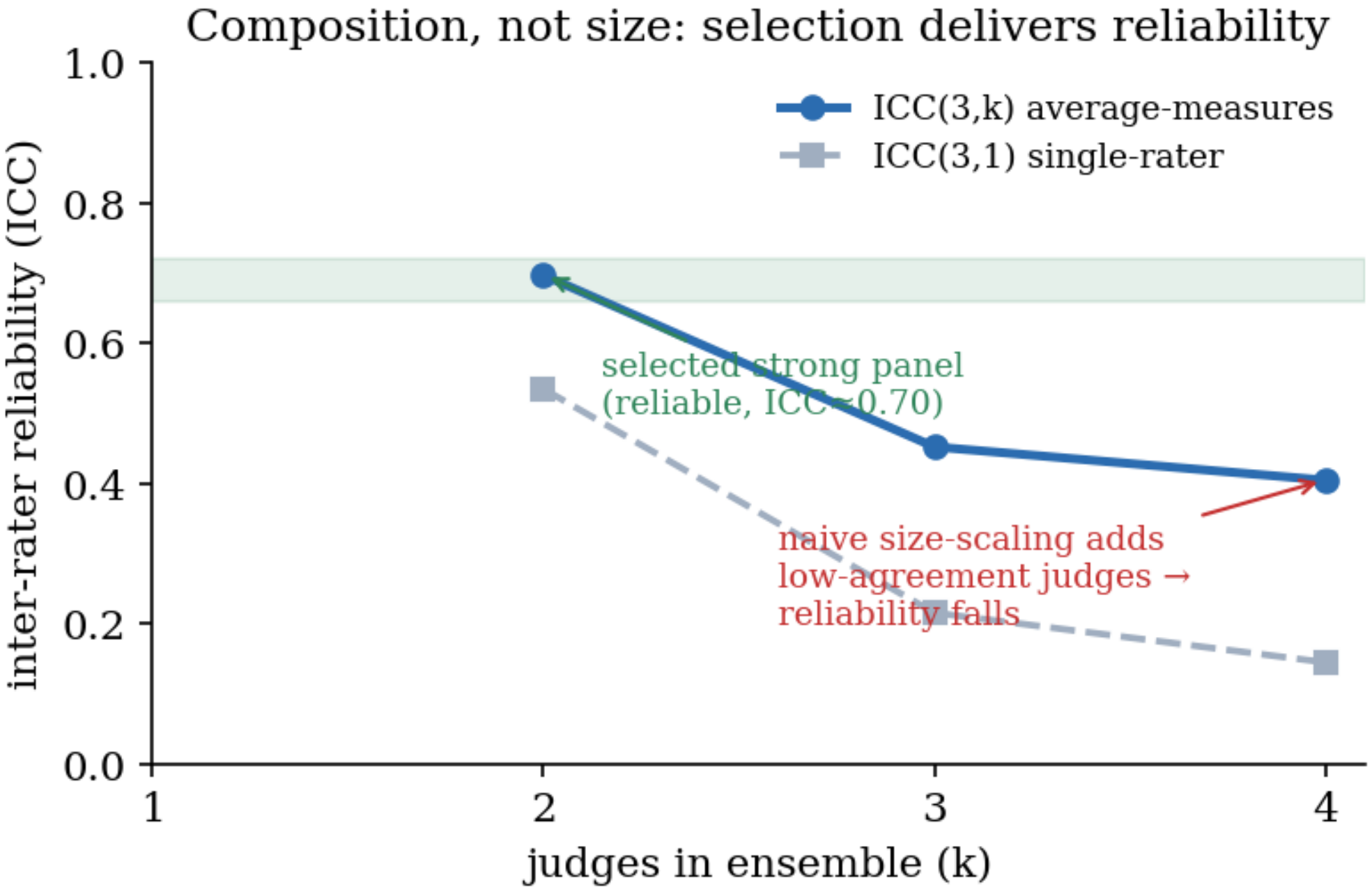

***Figure 3.*** *Composition over size. Average-measures reliability ICC(3, k) as judges are added in descending agreement order. A selected strong panel reaches ICC = 0.70; indiscriminately adding low-agreement judges lowers the mean inter-rater correlation faster than averaging compensates, so reliability falls. CoEval's consensus selection retains the high-agreement judges, delivering the reliable configuration automatically.*

Composition matters because the *wrong single judge can actively mislead*, and which judge is wrong is not knowable a priori without the labels CoEval assumes are absent. Across the benchmark-grounded set of Section 5.1 (three tasks, $n = 900$ pooled), the four candidate judges span a correlation range from $-0.04$ (gpt-3.5-turbo, anti-correlated) through 0.17 (gemini-flash) and 0.24 (gpt-4o-mini) to 0.31 (claude-haiku): a **judge-choice regret of** 0.35 between the best and worst single judge. The best judge is moreover task-dependent (claude-haiku here, gpt-4o-mini on the summarization subset of Section 5.1), so a practitioner committing to one judge in advance risks the anti-correlated one. The cross-family ensemble is the low-regret default: its plain mean correlates at 0.238 and, unlike a single pick, is *never* anti-correlated.

Given the panel, the choice of aggregator is itself a lever, and the unsupervised reliability-weighted aggregator of Section 3.3 is the best label-free option we test. Weighting each judge by its agreement with the rest of the panel assigns the anti-correlated gpt-3.5-turbo the lowest weight (0.18 versus 0.27–0.29 for the others) and lifts the aggregate correlation from the plain mean's 0.238 to 0.246. The alternatives are worse: median and trimmed-mean aggregation, which discard rather than re-weight information, reach only 0.222, and a Dawid–Skene latent-truth model [33] over discretized scores reaches 0.178. Reliability weighting thus turns the panel's own internal agreement, the one signal available without labels, into the strongest aggregate, while never selecting the anti-correlated judge.

Aggregation does more than average: it lets the evaluator *audit its own judges* with no labels. A judge's mean agreement with the rest of the panel is an unsupervised reliability signal, and it cleanly separates competent judges from broken ones. We stress-test this by adding three deliberately broken judges to the four-judge panel, one returning uniform-random scores, one a constant score, and one anti-correlated with quality. All three fall to the bottom of the agreement ranking (mean agreement $\leq 0$ versus 0.07 to 0.18 for the competent judges), and their effect on a naive average is large: the plain-mean correlation with ground truth drops from 0.238 to 0.126 once they are included. Reliability weighting, using only peer agreement, assigns each broken judge a weight of essentially zero and recovers the clean-panel accuracy (0.228). A single judge offers no such safeguard, because there is nothing to check it against: the panel turns the absence of labels from a liability into a design that validates itself.

This self-correction has a precise boundary, and the boundary is exactly the cross-family design. It tolerates any number of *independent* broken judges: adding six uniform-random judges to the four-judge panel leaves the reliability-weighted accuracy at 0.25 while the plain mean falls to 0.16, because independent noise never forms a consensus to agree with. It fails only when a *correlated* coalition of bad judges, all sharing one systematic error, *outnumbers* the competent judges: once five such judges face four good ones the coalition becomes the apparent consensus, takes all the weight, and the recovered ranking inverts. Peer-agreement weighting is therefore only as safe as the independence of judge errors, and *vendor diversity is what supplies that independence*: judges from distinct model families are unlikely to share a bias coalition. The robustness of the aggregator and the cross-family composition principle are two sides of one design.

The panel's two label-free quality signals, item discriminative power and judge agreement, compose into the *doubly-robust ranking* of Section 3.3, which leans on the items that separate models (Section 5.6) and the judges that agree with the panel. On a benchmark-grounded check where gold accuracy is available (thirteen models ranked on science QA and reasoning, spanning gpt-4o down to llama-3.2-1b), it improves rank-recovery of the true model ordering from the plain mean's Spearman 0.88 (Kendall 0.76) to 0.95 (Kendall 0.87), by down-weighting saturated items that carry no ranking signal (nearly half the items hold near-zero weight) and judges that disagree with the panel. The two signals are complementary, each helping on its own (item-weighting alone 0.94, judge-weighting alone 0.92) and most on the discriminative items where judge reliability matters. The aggregator is robust: injecting a random judge degrades the plain mean to 0.85 but leaves the doubly-robust ranking at 0.95, with the rogue judge receiving weight 0.00 (Table 4).

***Table 4.*** *Doubly-robust rank-recovery on thirteen candidate models (science QA and reasoning; gold = benchmark-native accuracy). Each label-free weight helps on its own, and the two compose into the best recovery; the aggregator is robust to an injected random judge, which it gives weight 0.00. Values are the Spearman and Kendall correlations between the recovered ranking and the gold-accuracy ranking.*

| Aggregator | Spearman | Kendall |
|---|---|---|
| Plain mean | 0.88 | 0.76 |
| Item-weighted (discrimination) only | 0.94 | 0.84 |
| Judge-weighted (agreement) only | 0.92 | 0.82 |
| Doubly-robust (both) | 0.95 | 0.87 |
| with injected random judge: plain mean | 0.85 | 0.71 |
| with injected random judge: doubly-robust | 0.95 | 0.87 |

### 5.3 Bias cancellation

We test whether CoEval's ensemble cancels verbosity bias, the tendency to reward longer responses, on the four-task medium-benchmark run, whose judge panel is gpt-4o-mini and gpt-3.5-turbo (OpenAI) together with qwen2.5-1.5b and smollm2-1.7b (open-weight). For each judge we compute the Pearson correlation between response length and assigned score. Individual judges carry length biases of *mixed sign*: gpt-3.5-turbo shows $r = -0.177$ (penalizing length), while smollm2-1.7b shows $r = +0.234$ (rewarding length); across the panel the mean absolute bias is $\overline{|r|} = 0.153$. No single judge is length-neutral.

The cancellation mechanism is the *mixed sign* of these biases under averaging, which a diverse panel supplies. We note a confound shared with self-preference (Section 5.4): in this panel, sign-diversity is correlated with both vendor family and model capability (the strong OpenAI judges penalize length; the small open-weight judges reward it), so we attribute the cancellation to bias-sign diversity rather than to vendor family *per se*. The practical recipe is the same: assemble judges whose idiosyncratic biases are unlikely to align, and vendor diversity is a convenient, observable proxy for it.

The **ensemble** correlation between length and score is $r = +0.010$, with a 95% bootstrap confidence interval of $[-0.039, +0.057]$ that *includes zero*, a 93% reduction in length-bias magnitude relative to the per-judge mean. The mixed-sign individual biases cancel under the averaging of $\bar{s}_i$, leaving an aggregate score that is statistically indistinguishable from length-neutral.

***Table 5.*** *Verbosity bias (Pearson r between response length and score) for representative individual judges versus the CoEval ensemble. The ensemble CI includes zero.*

| Judge | Length-bias $r$ | 95% bootstrap CI |
|---|---|---|
| gpt-3.5-turbo | −0.177 | – |
| SmolLM2 | +0.234 | – |
| Panel mean $\overline{\lvert r \rvert}$ | 0.153 | – |
| CoEval ensemble | +0.010 | [−0.039, +0.057] |

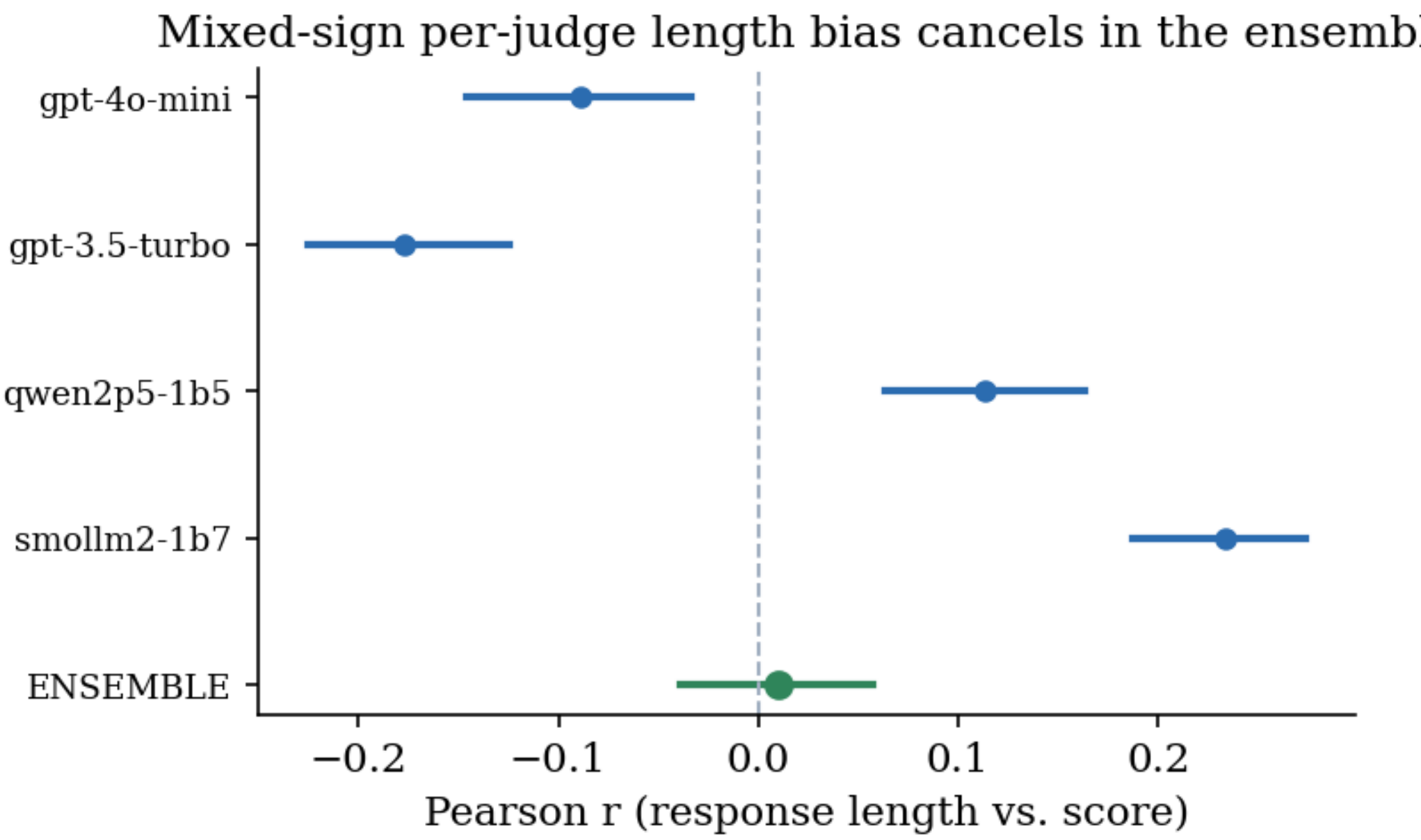

***Figure 4.*** *Verbosity bias by judge. Each judge's Pearson correlation between response length and score (point) with its 95% bootstrap CI; the ensemble (green) sits at r = +0.010 with a CI spanning zero. Mixed-sign per-judge biases cancel under averaging.*

### 5.4 Cross-family self-preference: a design property

Recent work establishes that judge–generator relatedness inflates scores: *Preference Leakage* [15] identifies same-model and same-family relatedness as a contamination vector, and *Play Favorites* [16] measures frontier models favoring same-family outputs. CoEval addresses this *structurally*: by requiring the judge panel to be vendor-disjoint from the student under test, it precludes a model from scoring its own family, eliminating the self-preference channel by design rather than by post-hoc correction. Because no judge ever scores its own family, a same-family preference cannot enter the aggregate in the first place, regardless of its magnitude on any individual model.

We also measure the residual same-family preference directly. The vertical case studies (Section 5.6) include gpt-4o-mini as both a judge and a candidate, which lets us isolate self-preference as a difference-in-differences that controls for a judge's overall harshness (how much more the same-family judge favors the in-family candidate over a rival than the cross-family judges do). Across the three verticals the residual effect is small and inconsistent in sign, +0.04 on clinical reasoning (95% CI $[-0.00, +0.08]$), −0.04 on legal, and −0.04 on drug-interaction: in a cross-family panel no systematic self-inflation survives, consistent with vendor diversity already neutralizing it. The structural guarantee makes this robustness automatic rather than incidental: because a model never scores its own family, a same-family preference of any sign or magnitude cannot enter the aggregate.

Beyond the structural guarantee, CoEval can *explicitly* account for systematic same-family bias through *vendor-disjoint aggregation*: when scoring a model under test, the ensemble drops any judge that shares the model's vendor family, so no model contributes to its own family's score. Applying this correction to the rankings of Tables 3 and 5 leaves *every* ranking unchanged, with per-model scores shifting by at most 0.016, a direct, auditable confirmation that the cross-family ensemble's rankings are already robust to same-family bias. The correction is therefore available as a safeguard, and its null effect here is itself evidence that vendor diversity, not a post-hoc adjustment, is doing the work.

### 5.5 Contamination resistance

We test the contamination-resistance claim empirically. We take the 400 CoEval-generated items from the medium-benchmark study and measure their verbatim 13-gram overlap against a corpus of 491 items drawn from five widely used public benchmarks (XSum, CNN/DailyMail, CodeSearchNet, SciQ, ARC-Challenge), datasets that are, by virtue of their age and popularity, present in the pretraining corpora of current models. Across 110,784 distinct public 13-grams, **not a single CoEval-generated item shares any 13-gram** with the public corpus: mean and maximum overlap are both 0.0000, and 0% of generated items share any 13-gram. This verbatim non-duplication *corroborates* the structural guarantee: because items are synthesized fresh from the attribute specification on each run, they cannot have appeared in any model's *prior* training, and a model cannot retrieve them from a leaked copy of these widely-used benchmarks. (The scope from Section 5.1 holds: this applies to CoEval's generated items; the benchmark-grounded anchor of Table 2 deliberately reuses public items to obtain objective ground truth.)

Non-duplication shows the items are fresh; a controlled study shows why that matters. We fine-tune a small open model (Qwen2.5-0.5B) to memorize 200 public SciQ items, turning that public set into a contaminated benchmark, and compare it against a clean frontier model (gpt-4o-mini) on both the memorized set and 100 fresh held-out items of the same capability. On the contaminated benchmark the memorizer scores a perfect 1.00 and is ranked *above* gpt-4o-mini (0.845); on the fresh items the order is correct, with gpt-4o-mini (0.81) above the memorizer (0.74, its fresh-measured capability). The memorized-minus-fresh gap isolates memorization from any skill the fine-tuning also imparted: the clean base model's gap is 0.10 (0.625 vs 0.53) while the contaminated model's is 0.26 (1.00 vs 0.74), so 0.16 of its apparent edge on the static benchmark is pure memorization that fresh items strip away. A static, possibly-contaminated leaderboard thus ranks a 0.5B memorizer ahead of a frontier model, an inversion that CoEval's freshly-generated items remove because no candidate can have trained on items synthesized after the fact.

### 5.6 Putting it together: domain case studies

The preceding sections validate CoEval's components where ground truth is available. We now exercise the complete tool in its intended regime on three custom verticals, *drug–drug interaction reasoning*, *clinical reasoning*, and *legal analysis*, for which we assume no task-specific labeled data and no trustworthy public benchmark. The drug-interaction vertical is the sharpest case for the framework's premise: a well-known public benchmark exists (DDIExtraction-2013) but is a relation-*extraction* corpus rather than clinical-decision reasoning, and is old and freely available enough to be presumed present in pretraining, while the few clinically realistic DDI-reasoning sets are private and number only in the hundreds of items. From a one-line description per vertical, CoEval's teacher generated 40 attribute-stratified items (for drug interactions, stratified over severity, mechanism, and patient context), three candidate models answered them, and the cross-family ensemble (OpenAI + Anthropic + Google) scored every response, all from a single configuration file, with no human labels or raters. Appendix C shows the actual generated artifacts for all three domains: the one-line seed, the auto-generated attribute strata and rubric, and a representative synthesized item.

On the drug-interaction vertical CoEval produces a **clean, fully unanimous ranking**: all three cross-family judges agree on the complete order gpt-4o-mini (0.770) > gpt-3.5-turbo (0.682) > llama-3.2-3b (0.497), with non-overlapping confidence intervals separating every model (Table 6). The clinical and legal verticals are consistent: the ensemble ranks the smallest model weakest in both, unanimously in clinical and two-of-three in legal (where the Anthropic judge ranks gpt-3.5-turbo lowest): genuine disagreement a single judge would hide and a diverse ensemble surfaces; on clinical the two stronger models are statistically close, which the overlapping intervals correctly expose. This is the operation the framework is built for: producing a defensible model ranking for a bespoke domain in which a standard benchmark is unavailable or untrustworthy.

A ranking is only possible if the items *separate* the models, the generation-side analog of judge discrimination: a good teacher writes questions on which candidates actually differ. We measure per-item discriminativeness as the spread of candidate-model ensemble scores. Across the 120 generated vertical items, 71% **are discriminative** (a score range ≥ 0.15 across the three models), with a mean per-item range of 0.33; on drug-interaction and legal analysis the figure rises to 78% and 85%. Where the items do not separate the models they are honest about it: clinical reasoning is only 50% discriminative precisely because its two stronger candidates are genuinely close (0.873 vs 0.864), so the items reflect a real near-tie rather than manufacturing a gap. This is why the panel yields non-overlapping ranking intervals: the teacher supplies items that carry ranking signal, and a non-discriminative benchmark, on which every model scores alike, could not separate the candidates no matter how reliable the judges are.

***Table 6.*** *CoEval ranks three candidate models on three custom, contamination-free verticals generated from a task description with no labeled data. Scores are the cross-family ensemble mean ([0, 1]) with 95% bootstrap CI over 40 generated items. llama-3.2-3b ranks weakest in every vertical; the three judges agree unanimously on the full drug-interaction order, unanimously on the weakest clinical model, and two-of-three in legal.*

| Vertical | Model | CoEval score [95% CI] |
|---|---|---|
| *Drug–drug interaction* | gpt-4o-mini | 0.770 [0.739, 0.800] |
| | gpt-3.5-turbo | 0.682 [0.646, 0.717] |
| | llama-3.2-3b | 0.497 [0.459, 0.532] |
| *Clinical reasoning* | gpt-3.5-turbo | 0.873 [0.849, 0.895] |
| | gpt-4o-mini | 0.864 [0.840, 0.885] |
| | llama-3.2-3b | 0.814 [0.784, 0.844] |
| *Legal analysis* | gpt-4o-mini | 0.982 [0.973, 0.991] |

| Vertical | Model | CoEval score [95% CI] |
|---|---|---|
| | gpt-3.5-turbo | 0.740 [0.709, 0.770] |
| | llama-3.2-3b | 0.709 [0.677, 0.744] |

### 5.7 Rankings are domain-specific: why a generic leaderboard misleads

The case studies above rank models *within* one domain. The premise of a custom-evaluation tool is that this is necessary: the best model for one domain need not be the best for another, so a single public leaderboard, which pools tasks into one number, can send a practitioner to the wrong model. We test this directly. From four one-line descriptions, *math word problems*, *code explanation*, *clinical reasoning*, and *legal analysis*, CoEval generated 25 contamination-free items each (100 items), six candidate models answered them, and the same vendor-disjoint cross-family panel (OpenAI + Anthropic + Google) scored every response: 1,800 evaluations at full coverage, from one configuration file.

The four per-domain rankings **disagree sharply**. Three different models take first place across the four domains, and each winner is stable under a bootstrap over items ($B = 2000$): gpt-4o-mini tops clinical reasoning (top with probability 0.88) and code explanation (0.60), gemini-flash tops legal analysis (1.00), and claude-3.5-haiku tops math word problems (0.85). The movement is not confined to the lead: gpt-4o-mini falls from first on clinical reasoning to fifth of six on math (where it is top in only 0.001 of bootstrap draws), and claude-3.5-haiku moves the opposite way, from first on math to fifth on code (Table 7). The mean cross-domain rank agreement is low (Kendall $\tau = 0.19$ averaged over the six domain pairs); the least-aligned pair, code explanation versus math word problems, has a negative point estimate ($\tau = -0.41$).

This is exactly the regret a generic leaderboard incurs. Pooling the four domains into one average ranks gemini-flash first, but that model is the correct pick for only one of the four domains (legal analysis); on clinical reasoning and code explanation the best model is gpt-4o-mini, and on math it is claude-3.5-haiku, which the pooled board ranks only third. A practitioner who reads the generic leaderboard and deploys its top model is choosing the domain-best model in just one of four cases. CoEval removes the guess: it ranks the candidates on the practitioner's own domain, contamination-free, with no labeled data.

***Table 7.*** *The same six candidate models ranked by the same cross-family panel on four CoEval-generated domains diverge: each cell is the model's rank (1 = best) in that domain, rows ordered by the generic pooled leaderboard. Three distinct models win the four domains (highlighted), and the pooled-best model (gemini-flash) is domain-best in only one. gpt-4o-mini swings from 1st to 5th and claude-3.5-haiku from 5th to 1st across domains.*

| Model | Pooled | Clinical | Code | Legal | Math |
|---|---|---|---|---|---|
| gemini-flash | 1 | 2 | 3 | 1 | 2 |
| gpt-4o-mini | 2 | 1 | 1 | 2 | 5 |
| claude-3.5-haiku | 3 | 3 | 5 | 3 | 1 |
| qwen-2.5-7b | 4 | 5 | 2 | 4 | 4 |
| llama-3.1-8b | 5 | 4 | 4 | 6 | 6 |
| gpt-3.5-turbo | 6 | 6 | 6 | 5 | 3 |

## 6. Discussion

CoEval is designed for a specific, common predicament: ranking models for a task or domain when no labeled data exists to build a benchmark and public benchmarks cannot be trusted. The results show it meets that need: its label-free scores reproduce the true model ranking and track ground truth, and its generated items are verifiably fresh. The composition-over-size finding explains *why* this label-free judging can be trusted. The key enabler is a design principle that is itself a contribution: the reliability of an LLM-judge panel is governed by its *composition*, not its size. A diverse, cross-family panel cancels the mixed-sign biases (verbosity, self-preference) that any individual judge carries, while a same-family panel would only amplify a shared bias; practitioners should prioritize vendor diversity over adding more copies of similar judges. A modest diverse panel already recovers most of the full-panel reliability, so the cost of diversity is small.

Coupling the panel to attribute-stratified generation is what makes the system more than a better judge. Because items are synthesized fresh from an attribute specification on every run, CoEval sidesteps contamination structurally rather than chasing leaked items after the fact, and it targets the deployment's own distribution and edge cases rather than a generic average. The same role separation that secures contamination-freeness in generation structurally precludes same-family self-preference in scoring: one architectural choice closing two leakage channels. Declarative specification keeps the whole pipeline reproducible: a single declarative configuration regenerates the benchmark, the responses, and the scores.

**Scope and limitations.** Our ground-truth validation of the ranking operation uses objective exact-match QA, where labels exist; on subjective custom domains (Section 5.6) we rely on the cross-family ensemble and report intervals rather than a held-out gold ranking, because by assumption none exists. We study up to thirteen candidate models across four exact-match QA tasks, three custom verticals, a four-domain divergence study (Section 5.7), and a thirteen-model rank-recovery check (Section 5.2); broader model pools and additional domains would further test generality. The contamination guarantee is structural (fresh per-run synthesis), validated both by zero verbatim overlap and by a controlled memorization study (Section 5.5); a verbatim n-gram test is not a membership test against any model's full pretraining corpus.

## 7. Conclusion

We presented CoEval, a framework for *ensemble self-evaluation* that ranks language models for a specific application when neither labeled data nor a trustworthy benchmark is available: from a task description alone, a pool of models rotates through all three roles, teacher, student, and judge, to generate a fresh, contamination-resistant benchmark, answer it, and rank the candidates, jointly weighting questions by discriminative power and judges by consensus, with no human annotation. Because every model also answers as a student, the same responses supply the data for both label-free weights. CoEval delivers bias-cancelled, reliable evaluation without human annotation: it removes verbosity bias that no single judge avoids (ensemble $r = +0.010$, CI spanning zero; 93% reduction), structurally precludes same-family self-preference, and produces task-specialized rubrics with a shared quality core, while also converging to a stable full-panel consensus (Spearman $\rho$ up to 1.00). Our central finding, that judge-panel *composition* rather than size is the decisive reliability lever, offers a concrete and inexpensive recipe for trustworthy, renewable LLM evaluation.

The panel is also self-validating: because a judge's agreement with its peers is a label-free reliability estimate, the ensemble detects and zeroes out broken judges with no ground truth, a safeguard a single judge cannot provide; the same weighting down-weights non-discriminative challenges, so saturated questions carry no spurious ranking signal. Because model rankings are *domain-specific* (three different models top four de-novo domains, so a generic leaderboard misdirects most practitioners) and the same declarative pipeline runs unchanged across models and domains, re-running on every model release, CoEval is a reusable instrument a team points at its own application when neither labeled data nor a trustworthy benchmark exists, rather than a single static study. CoEval is open-source and fully reproducible.

## Appendix A. Statistical methods

This appendix collects the estimator definitions used in the main text. CoEval reports the agreement of the judge panel using intraclass correlation. For an average-measures design over $k$ judges, the ICC(3,k) reliability is

$$\mathrm{ICC}(3,k) \; = \; \frac{MSR - MSE}{MSR},$$

where $MSR$ is the between-targets (rows) mean square and $MSE$ the residual mean square. The gain from aggregating more judges follows the **Spearman–Brown** prophecy relation: if $\bar{r}$ is the mean pairwise inter-judge correlation, the reliability of a $k$-judge mean is

$$R_k \; = \; \frac{k\,\bar{r}}{1 + (k-1)\,\bar{r}}.$$

For categorical agreement (e.g., pass/fail rubric criteria) CoEval reports Cohen's $\kappa$,

$$\kappa \; = \; \frac{p_o - p_e}{1 - p_e},$$

where $p_o$ is observed agreement and $p_e$ the agreement expected by chance. To quantify residual bias and its uncertainty, CoEval computes the Pearson correlation between a candidate confound (e.g., response length) and the assigned score, with a **nonparametric bootstrap** confidence interval: the per-item (length, score) pairs are resampled with replacement $B$ times, the correlation is recomputed on each resample, and the 2.5th–97.5th percentiles of the bootstrap distribution form the reported 95% CI. A CI that includes zero indicates that the corresponding bias is statistically indistinguishable from absent. Across the family of correlation tests reported in Section 5 we control the false-discovery rate with the Benjamini–Hochberg procedure ($\alpha = 0.05$), and confidence intervals use a *datapoint-clustered* resample wherever observations are nested (multiple student responses per item).

## Appendix B. Rubric structure

CoEval generates a scoring rubric automatically per task. We examine whether these rubrics are appropriately task-specialized while sharing a common quality core. Across the four tasks, the 22 auto-generated rubric criteria exhibit a mean *within-task* semantic similarity of 0.342, exceeding the mean *cross-task* similarity of 0.294: criteria cluster by task, as desired. At the same time, a shared universal *"completeness"* dimension recurs across three of the four tasks, indicating a common quality core. CoEval's rubrics are thus specialized to each task's demands yet anchored by a transferable notion of answer quality.

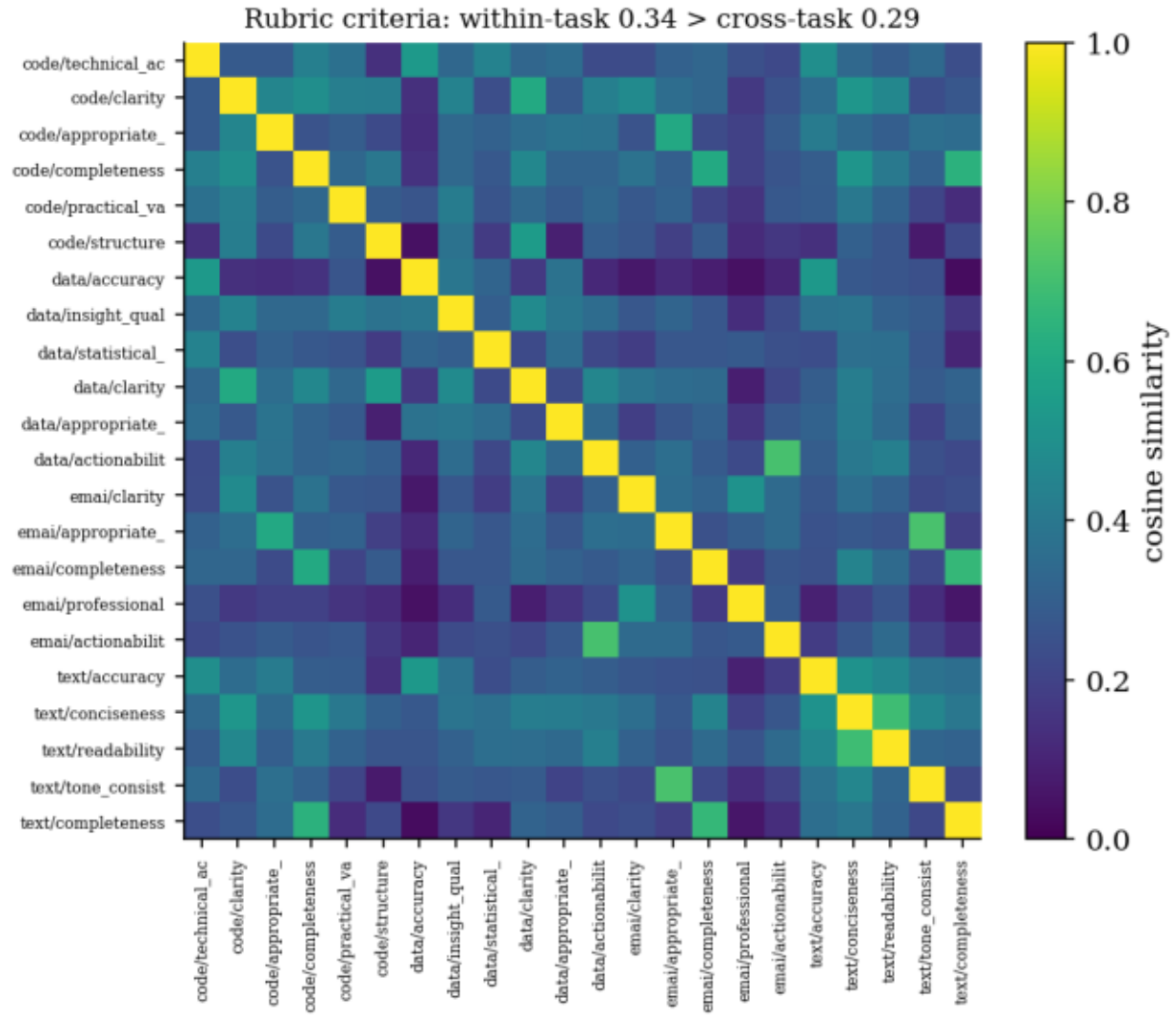

***Figure 5.*** *Pairwise semantic similarity of the 22 auto-generated rubric criteria across four tasks. Block structure along the diagonal reflects task specialization (within-task 0.342 > cross-task 0.294), while a recurring "completeness" dimension forms a cross-task cluster.*

## Appendix C. Worked examples on data-scarce domains

The three verticals of Section 5.6 are domains where a practitioner typically has *no* labeled benchmark and no trustworthy public one: drug–drug interaction (DDI) reasoning, clinical decision support, and legal analysis. The artifacts below are the *actual* output CoEval generated for each, from a single one-line task description and with no human labels. From that description the teacher inferred the attribute strata, synthesized 40 contamination-free items stratified over them, and wrote a scoring rubric; the vendor-disjoint panel (OpenAI + Anthropic + Google) then ranked three candidate models (Table 6).

The pipeline is four templated calls to the teacher, all instantiated from the one-line task_description (and an optional output_description). The first two build the attribute space the items are stratified over; the third writes the rubric the panel scores against; the fourth generates one contamination-free item per sampled attribute cell. The templates below are the verbatim framework prompts; a sampled cell such as *severity = moderate, mechanism = pharmacokinetic, patient_context = pregnancy* is what binds {target_attributes} and yields the corresponding DDI item shown after Table C1.

**(1) Target-attribute map.** Define the axes the benchmark varies over.

Generate synthetic data specifications for the {task_description} task by defining an attribute space that characterizes possible outputs. Return a JSON object mapping each attribute name to a list of possible values, and output only the JSON.

**(2) Nuance map.** Add surface-variation axes that change phrasing without changing the answer.

Define synthetic data specifications for the {task_description} task by creating a nuanced variability-focused attribute space. Include only attributes that change document phrasing, structure, noise, and context, without changing the underlying outputs. Return a single JSON object mapping each attribute name to a list of allowed values, and output only the JSON.

**(3) Auto-rubric.** Write the task-specific scoring factors the judges apply.

For the task {task_description}, where the model's output is {output_description}, create an evaluation rubric. Return only a JSON object where each key is a quality factor, and each value is a concise description of that factor. Output only the JSON.

**(4) Item synthesis.** Generate one realistic item for a sampled attribute cell.

Generate a realistic benchmark data point.
Task: {task_description}
Output format: {output_description}
Required attributes: {target_attributes}
Nuance: {nuanced_attributes}

Return only a JSON object with exactly two string keys:
"prompt": a realistic input text for the task
"response": the expected output for that input
Output only the JSON. No explanation or markdown.

Running this pipeline on the three one-line seeds below produces the auto-generated attribute strata and rubrics in Table C1, and the verbatim items that follow it.

***Table C1.*** *From a one-line description to a scored benchmark, for three domains with no public labeled data. The attribute strata and rubric criteria are auto-generated by the teacher; each domain yields 40 stratified, contamination-free items, scored by the cross-family panel.*

| Domain and one-line seed | Auto-generated attribute strata | Auto-generated rubric |
|---|---|---|
| **Drug–drug interaction reasoning** *"given two or more co-administered drugs and a patient context, assess the interaction, its mechanism, severity, and clinical action"* | severity {contraindicated, major, moderate, minor}; mechanism {pharmacokinetic, pharmacodynamic}; patient_context {renal, hepatic, polypharmacy-elderly, pregnancy} | interaction_accuracy, severity_correct, safety, completeness |

| Domain and one-line seed | Auto-generated attribute strata | Auto-generated rubric |
|---|---|---|
| **Clinical reasoning** *"a clinical reasoning question a clinician would face, requiring medical knowledge and safe, accurate reasoning"* | specialty {cardiology, infectious_disease, pediatrics, oncology, emergency}; difficulty {routine, intermediate, hard} | clinical_accuracy, safety, completeness, clarity |
| **Legal analysis** *"a legal analysis question requiring identification of the relevant rule and correct application to the facts"* | area_of_law {contracts, torts, criminal, constitutional, intellectual_property}; complexity {basic, intermediate, advanced} | legal_accuracy, reasoning_quality, completeness, clarity |

One representative *generated* item per domain (verbatim teacher output, labeled with its sampled attribute cell), showing the synthesized items are specific and realistic rather than templated:

5. **DDI** (severity = moderate, mechanism = pharmacokinetic, patient_context = pregnancy): *"A 30-year-old pregnant woman is prescribed lamotrigine for bipolar disorder and is also taking oral contraceptives. Assess the drug–drug interaction between these medications, identify the mechanism, and provide the severity and clinical recommendation."*
6. **Clinical** (specialty = cardiology, difficulty = routine): *"A 62-year-old male with a history of hypertension and hyperlipidemia presents with progressive shortness of breath and occasional palpitations. On examination, you note elevated jugular venous pressure and a third heart sound. An ECG shows left ventricular hypertrophy. What is the most likely diagnosis and what initial management strategy should be considered?"*
7. **Legal** (area_of_law = torts, complexity = basic): *"Alice, while jogging in the park, trips over a tree root that has been exposed due to erosion and falls, breaking her wrist. The park is owned by the city, and Alice claims the city is liable for her injuries due to negligence. Identify the relevant rule and determine if Alice can successfully hold the city liable."*

Each item is scored on its rubric by the panel, producing the rankings in Table 6. On DDI the three judges are unanimous, gpt-4o-mini (0.770) > gpt-3.5-turbo (0.682) > llama-3.2-3b (0.497), with non-overlapping confidence intervals; on clinical reasoning the two stronger models are statistically close (0.873 vs 0.864), which the overlapping intervals correctly expose. The entire path from a one-line description to a defensible, contamination-free ranking is a single configuration file with no labeled data and no human raters.